\documentclass{article}

\usepackage[utf8]{inputenc}
\usepackage[T1]{fontenc}
\usepackage{times}
\usepackage{helvet}
\usepackage{courier}

\usepackage{amsmath}
\usepackage{amssymb}
\usepackage{amsfonts}
\usepackage{amsthm}

\usepackage{booktabs}
\usepackage{multirow}
\usepackage{makecell}
\usepackage{array}
\usepackage{tabularx}

\usepackage{graphicx}
\usepackage{xcolor}
\usepackage{caption}

\usepackage{algorithm}
\usepackage{algpseudocode}

\usepackage{enumitem}
\usepackage{float}
\usepackage{nicefrac}
\usepackage{microtype}

\usepackage{url}
\usepackage[colorlinks=true, linkcolor=blue, citecolor=blue, urlcolor=blue]{hyperref}

\usepackage[numbers,square]{natbib}

\usepackage{arxiv}

\DeclareUnicodeCharacter{2013}{--}
\DeclareUnicodeCharacter{2014}{---}
\DeclareUnicodeCharacter{2018}{`}
\DeclareUnicodeCharacter{2019}{'}
\DeclareUnicodeCharacter{201C}{``}
\DeclareUnicodeCharacter{201D}{''}
\DeclareUnicodeCharacter{00D7}{$\times$}
\DeclareUnicodeCharacter{2191}{$\uparrow$}
\DeclareUnicodeCharacter{2193}{$\downarrow$}
\DeclareUnicodeCharacter{2265}{$\geq$}
\DeclareUnicodeCharacter{2264}{$\leq$}

\title{MotionCraft: Latent World Modeling with Sparse Attention for Visual Upscaling}

\author{
    Rong Fu \\
    Independent Researcher \\
    Corresponding author \and
    Chunlei Meng \\
    Independent Researcher \and
    Yangchen Zeng \\
    Independent Researcher \and
    Xiaowen Ma \\
    Independent Researcher \and
    Yongtai Liu \\
    Independent Researcher \and
    Wangyu Wu \\
    Independent Researcher \and
    Shuo Yin \\
    Independent Researcher \and
    Zijian Zhang \\
    Independent Researcher \and
    Sicheng Li \\
    Independent Researcher \and
    Yingrui Ji \\
    Independent Researcher \and
    Chenhao Wang \\
    Independent Researcher \and
    Simon Fong \\
    Independent Researcher
}

\renewcommand{\headeright}{}
\renewcommand{\undertitle}{}
\renewcommand{\shorttitle}{MotionCraft}

\hypersetup{
    pdftitle={MotionCraft: Latent World Modeling with Sparse Attention for Visual Upscaling},
    pdfsubject={cs.CV, cs.MM, cs.LG},
    pdfauthor={Rong Fu, Chunlei Meng, Yangchen Zeng, Xiaowen Ma, Yongtai Liu, Wangyu Wu, Shuo Yin, Zijian Zhang, Sicheng Li, Yingrui Ji, Chenhao Wang, Simon Fong},
    pdfkeywords={video super-resolution, visual upscaling, motion-aware modeling, sparse attention, latent world transformer, controllable video restoration},
    pdfstartview={FitH},
    colorlinks=true,
    linkcolor=red,
    citecolor=green,
    filecolor=magenta,
    urlcolor=cyan
}

\begin{document}
\maketitle

\begin{abstract}
Video super-resolution (VSR) aims to recover high-fidelity high-resolution videos from low-resolution inputs and is central to applications ranging from mobile capture to streaming and archival restoration. Existing approaches trade off among local-detail fidelity, long-range spatio-temporal modeling, perceptual realism and efficiency: convolutional alignment techniques preserve local structure but suffer when motion is large or degradations are complex; transformer-based methods capture long-range dependencies yet require architectural or algorithmic adaptations to be computationally feasible; and recent latent or diffusion-based generators synthesize rich texture but demand specialized temporal constraints to maintain coherence. In this work, we present \textbf{MotionCraft}, a controllable VSR framework that formulates restoration as motion-aware latent state prediction inspired by world models, and integrates adaptive sparse attention with an explicit, user-accessible control interface. MotionCraft combines robust motion fusion, a Latent World Transformer that balances locality and targeted non-local interactions, and a compact conditional decoder to deliver temporally consistent, high-quality reconstructions under streaming constraints. Empirical evaluations demonstrate that our design attains strong reconstruction and perceptual performance while enabling predictable trade-offs between temporal smoothness and fidelity. 
\end{abstract}


\section{Introduction}

Video super-resolution aims to recover high-quality, high-resolution video sequences from low-resolution inputs and remains a fundamental challenge in computer vision. This task finds application across diverse fields including mobile photography, media streaming, and the restoration of archived materials. Over the past decade, approaches have advanced from conventional spatio-temporal convolutional designs to architectures that harness extended correlations and robust generative priors. Recent survey articles and benchmarks underscore the growing body of research and the broad range of real-world scenarios where video super-resolution proves valuable \cite{baniya2024survey, xing2024survey}.

Contemporary methodologies in video super-resolution fall into three primary categories, each offering particular benefits yet also facing distinct constraints. Convolutional techniques equipped with alignment procedures excel at retaining local structures and crisp boundaries whenever displacements remain moderate and degradations stay controlled. Nevertheless, substantial movements or irregular degradation patterns cause misalignment errors to spread across the processing chain, impairing both image quality and temporal stability. In contrast, transformer architectures expand the receptive field considerably and facilitate the modeling of distant spatio-temporal relationships \cite{tang2023ctvsr, zhang2024RealViFormer}. Although capable of grasping intricate inter-frame linkages beyond what convolutional operators achieve, their standard full-attention implementation incurs high computational overhead for elevated resolutions or prolonged video lengths, thereby calling for sparse or block-based variants to ensure feasibility \cite{fuoli2023fast}. Generative methods operating in latent spaces, meanwhile, demonstrate exceptional proficiency in creating lifelike textures and perceptually appealing elements by drawing upon powerful learned priors within condensed representations \cite{blattmann2023align, zhou2024upscale, xu2025videogigagan}. These strategies expand opportunities for perceptual enhancement; however, the stochastic character of generation can produce discontinuities between frames unless supported by appropriate sequence conditioning, motion integration, or distillation procedures \cite{xing2024survey}.

Although notable strides have been made, three key obstacles hinder the effective application of high-quality video super-resolution in streaming and everyday contexts. Dense spatio-temporal attention mechanisms expand quadratically as spatial dimensions increase, rendering them unsuitable for high-resolution processing absent the adoption of sparse or block-sparse techniques \cite{fuoli2023fast, bahmani2025ac3d}. Moreover, dependence on external motion indicators such as optical flow or camera orientation data proves problematic in areas of occlusion, uniform texture, or intense activity, where direct incorporation of imprecise signals generates unwanted temporal inconsistencies \cite{jahedi2024ccmr, gehrig2021raft}. Additionally, latent generative frameworks supply compelling perceptual guidance but necessitate dedicated conditioning and distillation steps to safeguard precise adherence to reference information and to restrain random variations from one frame to the next \cite{xu2025videogigagan, chen2025dove}.

The proposed MotionCraft framework addresses these barriers by formulating restoration as motion-aware latent state prediction inspired by world models \cite{ding2025understanding}. It tightly couples motion reliability estimation, adaptive sparse attention, and latent-level user control into a single pipeline. Per-location confidence gating assesses external motion quality and actively conditions the sparse attention pattern, while a learned spatial gate modulates the latent state before decoding to enable smoothness and fidelity trade-offs. This yields an efficient, robust, and interactively controllable streaming system.

Our contributions are as follows. First, we unify motion-aware latent prediction with reliability-guided sparse attention, coupling confidence gating and block selection so that motion quality directly informs non-local retrieval. This differs from prior work that treats motion correction and attention in isolation; for instance, RealBasicVSR aligns flow without per-location reliability, RealViFormer attends block-wise without residual motion fusion, and FlashVSR caches latents without motion gating. Second, we design an adaptive sparse attention mechanism that preserves locality while retrieving remote blocks, jointly optimized with the motion branch. Third, we introduce a compact controllability interface that modulates the latent state before decoding via a global scalar and spatial gate. We provide a one-step streaming recipe with strong empirical performance.

\begin{figure}[t]
  \centering
  \resizebox{\textwidth}{!}{\includegraphics{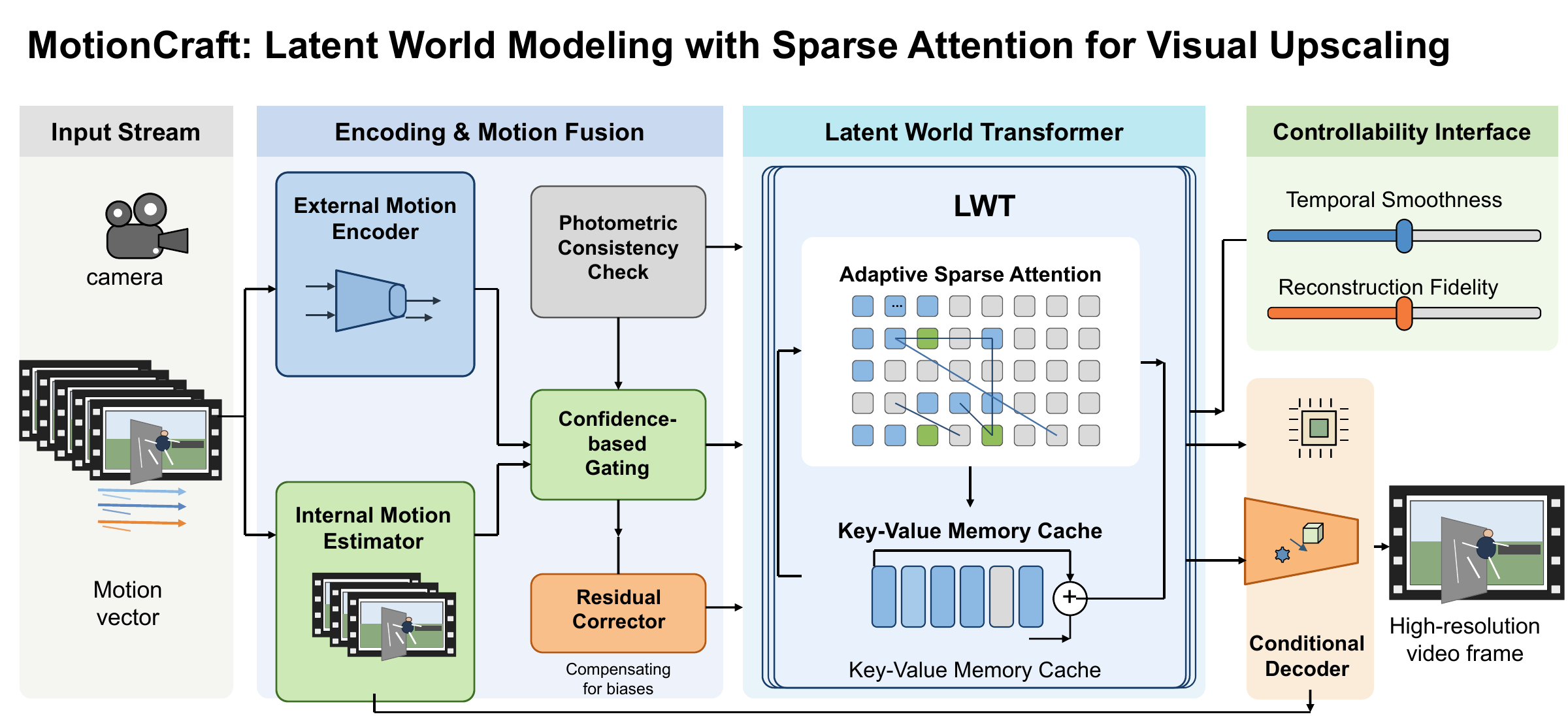}} 
  \caption{Overview of the \textbf{MotionCraft} framework for high-fidelity visual upscaling. 
  The system initiates with \textbf{Robust Motion Encoding and Fusion}, where external motion signals and internal image-driven estimates are gated by a \textbf{Photometric Inconsistency} confidence map to compensate for systematic biases. 
  The core \textbf{Latent World Transformer (LWT)} processes low-resolution latents using \textbf{Adaptive Sparse Attention}, which employs a differentiable soft top-$k$ selection to maintain locality while capturing long-range dependencies. 
  Temporal persistence is managed through a \textbf{Key-Value Memory Cache} ($\mathcal{M}_{t-1}$). 
  The \textbf{Controllability Interface} modulates the latent state via global scalars $\gamma$ and spatial gating maps $g_t$ to balance temporal smoothness and reconstruction fidelity. 
  Finally, a \textbf{Tiny Conditional Decoder} performs distillation-based refinement to produce the high-resolution output $\hat{X}_t$.} 
  \label{fig:MotionCraft_framework}
\end{figure}

\section{Related Work}
We position MotionCraft within four interrelated strands of prior research, including convolutional alignment methods, transformer-based spatio-temporal modeling and efficient attention, generative latent and diffusion approaches, and motion estimation combined with controllability mechanisms. The following subsections summarize representative work and highlight the gaps that MotionCraft addresses.
\subsection{Convolutional methods and explicit alignment}
Early VSR methods rely on explicit motion estimation and compensation to aggregate multi-frame evidence. Caballero et al.\cite{caballero2017real} proposed a spatio-temporal sub-pixel convolution pipeline for real-time VSR. Later work introduced implicit motion handling via dynamic filters and refinement to reduce flow sensitivity\cite{liu2020end}. RealBasicVSR\cite{chan2022investigating} addressed realistic degradations with pre-cleaning and stochastic strategies. Despite improvements, pixel-level alignment errors still degrade temporal coherence under large motion or severe artifacts\cite{isobe2020video}.

\subsection{Transformer-based spatio-temporal modeling and attention efficiency}
Transformers offer a natural mechanism for long-range spatio-temporal modeling. Trajectory-aware Transformer (TTVSR) restricts attention to pre-aligned token trajectories, enabling long-range modeling with reduced cost\cite{liu2022learning}. Collaborative transformer designs combine multi-scale spatial tokens with temporal trajectories to balance throughput and accuracy\cite{tang2023ctvsr}. To tame full attention costs, recent works adopt adaptive sparse selection, block-wise retrieval, or deformable attention to preserve locality while allowing targeted non-local interactions\cite{geng2022rstt,fuoli2023fast}. The role of alignment in transformer pipelines has been revisited: some studies find that unaligned inputs can be directly exploited by attention mechanisms, and propose patch-level alignment as a computationally attractive compromise\cite{shi2022rethinking,zhang2023multi}.

\subsection{Diffusion and latent-space generative methods}
Latent diffusion and cascaded strategies have recently been adapted for VSR to leverage strong generative priors while limiting compute in pixel space\cite{blattmann2023align,xie2025simplegvr}. Text-guided and motion-guided latent diffusion frameworks demonstrate that controlled sampling and temporal modules can improve perceptual quality on real-world videos\cite{zhou2024upscale,yang2024motion}. Other lines of work argue that a powerful space-time diffusion transformer can implicitly model motion priors and reduce reliance on explicit flow alignment\cite{zhan2025rethinking,li2025diffvsr}. Despite their advantages for texture synthesis, diffusion-based VSR methods must address randomness and temporal discontinuities, typically through motion-guided losses or sequence-oriented fine-tuning\cite{zhang2025flashvideo}.

\subsection{Motion estimation and controllability}
Progress in dense optical flow and high-resolution flow estimation benefits both alignment-based and hybrid pipelines\cite{gehrig2021raft,jahedi2024ccmr,xu2021high}. For applications requiring user interaction, temporal modulation and controllable interpolation schemes enable adjustable trade-offs between fidelity and smoothness\cite{xu2021temporal}. In MotionCraft we explicitly estimate per-location motion reliability, learn a compact residual to correct biased external motion cues, and expose a combined global scalar and spatial gate for fine-grained, user-level control. This design yields predictable control of motion smoothness versus reconstruction fidelity while maintaining temporal stability. 

\section{Methodology}
\label{sec:method}

\subsection{Overview}
MotionCraft formulates video super-resolution as controlled state prediction in a compact latent space. At time $t$, the model processes the current low-resolution frame, a bounded history of observed frames, and motion fields computed only from frame pairs that have already arrived. External and image-driven motion representations are combined through spatial reliability estimation and residual correction. A Latent World Transformer then predicts the high-resolution latent using adaptive sparse attention and a causal key-value cache. A compact control interface regulates the balance between temporal smoothness and reconstruction fidelity, while a Tiny Conditional Decoder reconstructs the output frame. Future frames may provide auxiliary supervision during offline training, but they are never supplied to the inference network that produces the prediction at time $t$.

\subsection{Formal Problem Formulation}
Let the low-resolution input sequence and its high-resolution reference be denoted by $\mathcal{I}=\{I_t\}_{t=1}^{T}$ and $\mathcal{X}=\{X_t\}_{t=1}^{T}$, respectively. For a causal history radius $\tau$, the first retained index and the available frame history are defined as
\begin{equation}
 s_t=\max(1,t-\tau),
 \qquad
 \mathcal{H}^{I}_t=\{I_i\}_{i=s_t}^{t}.
 \label{eq:causal_frame_history}
\end{equation}
where $s_t$ truncates the temporal window near the beginning of a sequence, $\tau$ is the maximum history length, and $\mathcal{H}^{I}_t$ contains only frames observed no later than time $t$.

The corresponding causal motion history is
\begin{equation}
 \mathcal{H}^{f}_t
 =
 \left\{f_{i\rightarrow i+1}\right\}_{i=s_t}^{t-1}.
 \label{eq:causal_flow_history}
\end{equation}
where $f_{i\rightarrow i+1}$ is estimated from the already available pair $(I_i,I_{i+1})$, and $\mathcal{H}^{f}_1$ is represented by a learned null-motion token.

The streaming prediction is expressed as
\begin{equation}
 \hat X_t
 =
 F_{\Theta}\!\left(
 \mathcal{H}^{I}_t,
 \mathcal{H}^{f}_t,
 \mathcal{M}_{t-1};
 \gamma,g_t
 \right).
 \label{eq:causal_prediction}
\end{equation}
where $F_{\Theta}$ is the complete MotionCraft network with parameters $\Theta$, $\mathcal{M}_{t-1}$ is the key-value memory constructed from previous frames, $\gamma$ is a global user control, and $g_t$ is a learned spatial control map.

Strict causality requires
\begin{equation}
 \hat X_t
 \perp\!\!\!\perp
 \{I_j\}_{j=t+1}^{T}
 \;\mid\;
 \left(\mathcal{H}^{I}_t,\mathcal{H}^{f}_t,\mathcal{M}_{t-1}\right).
 \label{eq:causal_independence}
\end{equation}
where the conditional-independence relation states that future observations cannot affect the output emitted for time $t$.

\subsection{Causal Motion Encoding and Reliability-Guided Fusion}
External motion cues are aggregated over the causal history:
\begin{equation}
 m^{\mathrm{ext}}_t
 =
 \mathcal{F}_{\mathrm{ext}}\!\left(
 \left\{f_{i\rightarrow i+1}\right\}_{i=s_t}^{t-1}
 \right).
 \label{eq:m_ext_rewrite}
\end{equation}
where $m^{\mathrm{ext}}_t$ is the external motion embedding, $\mathcal{F}_{\mathrm{ext}}$ is a lightweight motion encoder, and the latest admissible field is $f_{t-1\rightarrow t}$.

The internal branch extracts motion evidence directly from the observed image history:
\begin{equation}
 m^{\mathrm{int}}_t
 =
 \mathcal{F}_{\mathrm{int}}\!\left(I_{s_t:t}\right).
 \label{eq:m_int_rewrite}
\end{equation}
where $m^{\mathrm{int}}_t$ is the internal motion embedding, $\mathcal{F}_{\mathrm{int}}$ is a compact reduced-resolution encoder, and $I_{s_t:t}$ denotes the ordered frame buffer from $I_{s_t}$ to $I_t$.

For $t>1$, the reliability of the external estimate is evaluated by causal bidirectional photometric inconsistency:
\begin{align}
 e_t
 ={}&
 \mathrm{PhotErr}\!\left(
 I_t,
 \mathrm{Warp}\!\left(I_{t-1},f_{t-1\rightarrow t}\right)
 \right)
 \nonumber\\
 &+
 \mathrm{PhotErr}\!\left(
 I_{t-1},
 \mathrm{Warp}\!\left(I_t,f_{t\rightarrow t-1}\right)
 \right).
 \label{eq:phot_err_rewrite}
\end{align}
where $e_t$ is a per-location inconsistency map, $f_{t-1\rightarrow t}$ and $f_{t\rightarrow t-1}$ are both estimated from $(I_{t-1},I_t)$, $\mathrm{Warp}$ is differentiable motion compensation, and $\mathrm{PhotErr}$ is a robust photometric discrepancy. For $t=1$, $e_t$ is initialized to zero.

A compact calibration head converts this inconsistency into a spatial confidence map:
\begin{equation}
 c_t
 =
 \sigma\!\left(
 \mathcal{C}_{\omega}\!\left(
 \left[\mathrm{Pool}(e_t),\mathrm{feat}_t\right]
 \right)
 \right).
 \label{eq:conf_rewrite}
\end{equation}
where $c_t\in[0,1]$ measures the local reliability of the external branch, $\mathcal{C}_{\omega}$ is a shallow convolutional head, $\mathrm{Pool}$ performs coarse spatial aggregation, $\mathrm{feat}_t$ contains lightweight auxiliary features, $[\cdot,\cdot]$ denotes channel concatenation, and $\sigma$ is the sigmoid function.

A causal residual corrector compensates for systematic motion bias:
\begin{equation}
 r_t
 =
 \mathcal{R}_{\rho}\!\left(
 I_{s_t:t},m^{\mathrm{ext}}_t
 \right).
 \label{eq:residual_rewrite}
\end{equation}
where $r_t$ is the predicted correction and $\mathcal{R}_{\rho}$ is a compact residual network parameterized by $\rho$.

The fused motion representation is
\begin{equation}
 a_t
 =
 \mathcal{M}_{\phi}\!\left(
 c_t\odot m^{\mathrm{ext}}_t
 +(1-c_t)\odot m^{\mathrm{int}}_t
 +r_t
 \right).
 \label{eq:motion_fuse_rewrite}
\end{equation}
where $a_t$ is the fused embedding, $\mathcal{M}_{\phi}$ is a projection network with parameters $\phi$, and $\odot$ denotes element-wise multiplication with spatial broadcasting when necessary.

\subsection{Latent World Transformer with Adaptive Sparse Attention}
The current frame is encoded into a compact latent and transformed into a high-resolution latent state:
\begin{align}
 z^L_t
 & =
 \mathcal{E}_{\eta}(I_t),
 \nonumber\\
 \hat z^H_t
 & =
 \mathcal{T}_{\theta}\!\left(
 z^L_t,\tilde a_t,\mathcal{M}_{t-1}
 \right).
 \label{eq:lwt_rewrite}
\end{align}
where $\mathcal{E}_{\eta}$ is the low-resolution encoder, $z^L_t$ is the current latent, $\mathcal{T}_{\theta}$ is the Latent World Transformer, $\tilde a_t$ is the controlled motion embedding defined below, and $\mathcal{M}_{t-1}$ contains key-value summaries from earlier frames only. Inspired by world models, this formulation treats video restoration as sequential state prediction in a compact latent space, but it does not instantiate a full action-environment interaction loop. To preserve locality while limiting attention cost, spatial tokens are partitioned into blocks and pooled into query-key summaries:
\begin{equation}
 \bar q_b
 =
 \mathrm{Pool}\!\left(\{q_i:i\in b\}\right),
 \qquad
 \bar k_b
 =
 \mathrm{Pool}\!\left(\{k_j:j\in b\}\right).
 \label{eq:block_pool_rewrite}
\end{equation}
where $b$ indexes a spatial block, $q_i$ and $k_j$ are token-level query and key vectors, and $\bar q_b$ and $\bar k_b$ are the corresponding block descriptors.

The relevance of candidate block $b$ to query block $b_q$ is scored by
\begin{equation}
 s_{b_q,b}
 =
 \frac{\left\langle\bar q_{b_q},\bar k_b\right\rangle}{\sqrt{d}}.
 \label{eq:block_score_rewrite}
\end{equation}
where $s_{b_q,b}$ is the block relevance score, $\langle\cdot,\cdot\rangle$ denotes the inner product, and $d$ is the attention dimensionality. During training, a differentiable soft top-$k$ relaxation assigns continuous weights to candidate blocks. During inference, hard top-$k$ retrieval restricts full token attention to the local neighborhood and the highest-scoring remote blocks. Temporal masking prevents the current state from attending to frames after time $t$.

\subsection{Controllability Interface and Conditional Decoding}
The fused motion state is modulated by a global coefficient and a spatial gate:
\begin{equation}
 \tilde a_t(\gamma,g_t)
 =
 \left(\gamma\mathbf{1}+g_t\right)\odot a_t.
 \label{eq:control_rewrite}
\end{equation}
where $\tilde a_t$ is the controlled motion embedding, $\gamma$ is a user-specified scalar, $\mathbf{1}$ is an all-ones tensor compatible with $a_t$, and $g_t$ is a learned spatial gate regularized toward small magnitude. The Tiny Conditional Decoder reconstructs the high-resolution frame as
\begin{equation}
 \hat X_t
 =
 \mathcal{D}_{\psi}\!\left(\hat z^H_t,I_t\right).
 \label{eq:decoder_rewrite}
\end{equation}
where $\mathcal{D}_{\psi}$ is the decoder parameterized by $\psi$, $\hat z^H_t$ supplies predicted high-frequency content, and $I_t$ preserves spatially aligned low-frequency structure.

Decoder distillation transfers fine detail from a stronger teacher:
\begin{align}
 \mathcal{L}_{\mathrm{dec\_distill}}
 ={}&
 \alpha_{\mathrm{feat}}
 \left\lVert
 \Phi(\hat X_t)-\Phi(X^{\mathrm{teach}}_t)
 \right\rVert_2^2
 \nonumber\\
 &+
 \alpha_{\mathrm{pix}}
 \left\lVert
 \hat X_t-X^{\mathrm{teach}}_t
 \right\rVert_2^2.
 \label{eq:dec_distill_rewrite}
\end{align}
where $X^{\mathrm{teach}}_t$ is the teacher reconstruction, $\Phi$ is a fixed perceptual feature extractor, and $\alpha_{\mathrm{feat}}$ and $\alpha_{\mathrm{pix}}$ balance feature-space and pixel-space supervision.

\subsection{Training Objectives}
The JEPA-style representation objective is
\begin{equation}
 \mathcal{L}_{\mathrm{JEPA}}
 =
 \mathbb{E}\!\left[
 \left\lVert\hat\ell_t-\ell_t\right\rVert_2^2
 \right].
 \label{eq:jepa_rewrite}
\end{equation}
where $\hat\ell_t$ is the student prediction and $\ell_t$ is the target representation produced by a momentum encoder.

Temporal coherence is supervised by a one-step latent warping loss:
\begin{equation}
 \mathcal{L}_{\mathrm{flow}}
 =
 \mathbb{E}\!\left[
 \left\lVert
 \mathrm{Warp}\!\left(\hat z^H_t,f_{t\rightarrow t+1}\right)
 -\hat z^H_{t+1}
 \right\rVert_1
 \right].
 \label{eq:flow_loss_rewrite}
\end{equation}
where $f_{t\rightarrow t+1}$ and $\hat z^H_{t+1}$ are available from the complete training clip and are used only to evaluate the loss. Neither quantity is provided as an input when the student predicts $\hat z^H_t$. Teacher and student latents are aligned by paired latent regression:
\begin{equation}
 \mathcal{L}_{\mathrm{latent\_distill}}
 =
 \mathbb{E}\!\left[
 \left\lVert
 z^{\mathrm{teach}}_t-z^{\mathrm{stud}}_t
 \right\rVert_2^2
 \right].
 \label{eq:latent_distill_rewrite}
\end{equation}
where $z^{\mathrm{teach}}_t$ and $z^{\mathrm{stud}}_t$ are paired teacher and student latents for the same frame. This objective is termed latent distillation because it performs sample-wise regression rather than explicit distribution matching.

The complete training objective is
\begin{align}
 \mathcal{L}_{\mathrm{total}}
 ={}&
 \lambda_{\mathrm{latent}}\mathcal{L}_{\mathrm{latent\_distill}}
 +\lambda_{\mathrm{flow}}\mathcal{L}_{\mathrm{flow}}
 +\lambda_{\ell_2}\mathcal{L}_{\ell_2}
 +\lambda_{\mathrm{LPIPS}}\mathcal{L}_{\mathrm{LPIPS}}
 \nonumber\\
 &+
 \lambda_{\mathrm{JEPA}}\mathcal{L}_{\mathrm{JEPA}}
 +\lambda_{\mathrm{adv}}\mathcal{L}_{\mathrm{adv}}
 +\lambda_{\mathrm{conf}}\mathcal{L}_{\mathrm{conf}}
 +\lambda_{\mathrm{dec}}\mathcal{L}_{\mathrm{dec\_distill}}.
 \label{eq:total_loss_rewrite}
\end{align}
where each $\lambda_{\bullet}\geq0$ controls the contribution of its associated loss, $\mathcal{L}_{\ell_2}$ is the reconstruction loss, $\mathcal{L}_{\mathrm{LPIPS}}$ is the perceptual loss, $\mathcal{L}_{\mathrm{adv}}$ is the adversarial loss, and $\mathcal{L}_{\mathrm{conf}}$ calibrates motion reliability.

\subsection{Unified Training and Strictly Causal Streaming Procedure}
During offline training, complete clips may be used to construct teacher targets and the one-step temporal loss. The student input at time $t$ remains restricted to $I_{s_t:t}$, $\mathcal{H}^{f}_t$, and $\mathcal{M}_{t-1}$. At test time, the prediction is emitted before $I_{t+1}$ arrives. This execution order has zero future-frame look-ahead. End-to-end latency is measured from the arrival of $I_t$ to the emission of $\hat X_t$ and includes bidirectional motion estimation, motion fusion, LWT inference, conditional decoding, and cache update. Consequently, FPS measures processing throughput rather than per-frame response latency. 

\begin{algorithm}[t]
\caption{MotionCraft training and strictly causal streaming inference}
\label{alg:motioncraft}
\begin{algorithmic}[1]
\Require Image-video training data, optional teacher, history radius $\tau$, external motion estimator, controls $(\gamma,g_t)$, stage flag
\Ensure Trained parameters and emitted frames $\{\hat X_t\}$
\If{stage $=$ Full}
  \State Train a dense teacher on image and video data
  \State Introduce sparse attention and causal temporal masks
  \State Distill the teacher into the one-step LWT student using Eq.~\eqref{eq:total_loss_rewrite}
\Else
  \State Train the sparse causal student directly with reconstruction, latent, and temporal supervision
\EndIf
\State Initialize the causal frame buffer, motion buffer, and key-value cache $\mathcal{M}_0$
\For{each arriving frame $I_t$}
  \State Set $s_t\leftarrow\max(1,t-\tau)$ and append $I_t$ to $I_{s_t:t}$
  \State Encode $I_t$ to obtain $z^L_t\leftarrow\mathcal{E}_{\eta}(I_t)$
  \If{$t>1$}
    \State Estimate $f_{t-1\rightarrow t}$ and $f_{t\rightarrow t-1}$ from $(I_{t-1},I_t)$
    \State Update $\mathcal{H}^{f}_t\leftarrow\{f_{i\rightarrow i+1}\}_{i=s_t}^{t-1}$ and compute $e_t$
  \Else
    \State Use the null-motion token and set $e_t\leftarrow0$
  \EndIf
  \State Compute $m^{\mathrm{ext}}_t$ from $\mathcal{H}^{f}_t$ and $m^{\mathrm{int}}_t$ from $I_{s_t:t}$
  \State Compute $c_t$, $r_t$, and $a_t$ using Eqs.~\eqref{eq:conf_rewrite}--\eqref{eq:motion_fuse_rewrite}
  \State Apply the control interface to obtain $\tilde a_t(\gamma,g_t)$
  \State Predict $\hat z^H_t\leftarrow\mathcal{T}_{\theta}(z^L_t,\tilde a_t,\mathcal{M}_{t-1})$
  \State Decode $\hat X_t\leftarrow\mathcal{D}_{\psi}(\hat z^H_t,I_t)$
  \State Update the causal cache $\mathcal{M}_t$ and emit $\hat X_t$ before receiving $I_{t+1}$
\EndFor
\end{algorithmic}
\end{algorithm}

\begin{table}[t]
\centering
\caption{Quantitative comparison with state-of-the-art video super-resolution methods. 
Our method, MotionCraft, demonstrates significant advantages across all datasets and metrics. Complementary quantitative comparison on UDM10 and MovieLQ under
the evaluation protocol of InfVSR~\cite{zhang2025infvsr}.}
\label{tab:quantitative_comparison_infvsr}
\resizebox{\textwidth}{!}{%
\begin{tabular}{@{}llccccccccc@{}}
\toprule
\textbf{Dataset} & \textbf{Metric}
& \textbf{RealBasicVSR\cite{chan2022investigating}}
& \textbf{RealViFormer\cite{zhang2024RealViFormer}}
& \textbf{Upscale-A-Video\cite{zhou2024upscale}}
& \textbf{MGLD-VSR\cite{yang2024motion}}
& \textbf{STAR\cite{xie2025star}}
& \textbf{SeedVR\cite{wang2025seedvr}}
& \textbf{SeedVR2\cite{wang2025seedvr2}}
& \textbf{InfVSR\cite{zhang2025infvsr}}
& \textbf{MotionCraft (Ours)} \\
\midrule
\multirow{8}{*}{UDM10\cite{tao2017detail}}
& PSNR $\uparrow$
& 24.13 & 24.64 & 21.72 & 24.23 & 23.47 & 23.39 & 25.38 & 24.86
& \textbf{28.15} \\
& SSIM $\uparrow$
& 0.6801 & 0.6947 & 0.5913 & 0.6957 & 0.6804 & 0.6843 & 0.7764 & 0.7274
& \textbf{0.8200} \\
& LPIPS $\downarrow$
& 0.3908 & 0.3681 & 0.4116 & 0.3272 & 0.4242 & 0.3583 & 0.2868 & 0.2972
& \textbf{0.2450} \\
& DISTS $\downarrow$
& 0.2067 & 0.2039 & 0.2230 & 0.1677 & 0.2156 & 0.1339 & 0.1512 & 0.1422
& \textbf{0.1180} \\
& MUSIQ $\uparrow$
& 59.06 & 57.90 & 59.91 & 60.55 & 41.98 & 53.62 & 49.95 & 62.88
& \textbf{76.30} \\
& CLIP-IQA $\uparrow$
& 0.3494 & 0.4157 & 0.4697 & 0.4557 & 0.2417 & 0.3145 & 0.2987 & 0.5142
& \textbf{0.6550} \\
& DOVER $\uparrow$
& 0.7564 & 0.7303 & 0.7291 & 0.7264 & 0.4830 & 0.6889 & 0.5568 & 0.7826
& \textbf{0.8720} \\
& $E^{*}_{\mathrm{warp}}~(\times 10^{-3}) \downarrow$
& 3.10 & 2.29 & 3.97 & 3.59 & 2.08 & 3.24 & 1.98 & 1.95
& \textbf{1.62} \\
\midrule
\multirow{4}{*}{MovieLQ\cite{zhang2025infvsr}}
& MUSIQ $\uparrow$
& 62.59 & 63.74 & 68.49 & 67.90 & 56.57 & 64.42 & 61.13 & 68.65
& \textbf{72.80} \\
& CLIP-IQA $\uparrow$
& 0.4672 & 0.4227 & 0.5117 & 0.5591 & 0.3411 & 0.5050 & 0.4468 & 0.5888
& \textbf{0.6350} \\
& DOVER $\uparrow$
& 0.8234 & 0.8273 & 0.7750 & 0.8402 & 0.7565 & 0.8145 & 0.8031 & 0.8447
& \textbf{0.8800} \\
& $E^{*}_{\mathrm{warp}}~(\times 10^{-3}) \downarrow$
& 3.39 & 2.24 & 5.53 & 3.67 & 3.11 & 4.70 & 4.26 & 2.88
& \textbf{1.95} \\
\bottomrule
\end{tabular}%
}
\end{table}

\begin{table}[t]
\centering
\caption{Quantitative comparison of video super-resolution methods across multiple datasets. Best results are in \textbf{bold}. Quantitative comparison under the evaluation protocol of
FlashVSR~\cite{zhuang2025flashvsr}.}
\label{tab:quantitative_comparison_flashvsr}

\resizebox{\textwidth}{!}{%
\begin{tabular}{@{}llcccccccc@{}}
\toprule
\textbf{Dataset} & \textbf{Metric}
& \textbf{Upscale-A-Video\cite{zhou2024upscale}}
& \textbf{STAR\cite{xie2025star}}
& \textbf{RealViFormer\cite{zhang2024RealViFormer}}
& \textbf{DOVE\cite{chen2025dove}}
& \textbf{SeedVR2-3B\cite{wang2025seedvr2}}
& \textbf{FlashVSR-Full\cite{zhuang2025flashvsr}}
& \textbf{FlashVSR-Tiny\cite{zhuang2025flashvsr}}
& \textbf{MotionCraft (Ours)} \\
\midrule
\multirow{7}{*}{YouHQ40\cite{zhou2024upscale}}
& PSNR $\uparrow$
& 23.19 & 23.19 & 23.67 & 24.39 & 23.05 & 23.13 & 23.31
& \textbf{25.42} \\
& SSIM $\uparrow$
& 0.6075 & 0.6388 & 0.6189 & 0.6651 & 0.6248 & 0.6004 & 0.6110
& \textbf{0.7180} \\
& LPIPS $\downarrow$
& 0.4585 & 0.4705 & 0.4476 & 0.4011 & 0.3876 & 0.3874 & 0.3866
& \textbf{0.3620} \\
& NIQE $\downarrow$
& 4.834 & 7.275 & 3.360 & 4.890 & 3.751 & 3.382 & 3.489
& \textbf{3.082} \\
& MUSIQ $\uparrow$
& 43.07 & 35.05 & 62.73 & 61.60 & 62.31 & 69.16 & 66.63
& \textbf{73.25} \\
& CLIP-IQA $\uparrow$
& 0.3380 & 0.2974 & 0.4451 & 0.4437 & 0.4909 & 0.5873 & 0.5221
& \textbf{0.6310} \\
& DOVER $\uparrow$
& 6.889 & 7.363 & 9.739 & 11.29 & 12.43 & 12.71 & 12.66
& \textbf{13.28} \\
\midrule
\multirow{7}{*}{REDS\cite{nah2019ntire}}
& PSNR $\uparrow$
& 24.84 & 24.01 & 25.96 & 25.60 & 24.83 & 23.92 & 24.11
& \textbf{27.05} \\
& SSIM $\uparrow$
& 0.6437 & 0.6765 & 0.7092 & 0.7257 & 0.7042 & 0.6491 & 0.6511
& \textbf{0.8050} \\
& LPIPS $\downarrow$
& 0.4168 & 0.3710 & 0.2997 & 0.3077 & 0.3124 & 0.3439 & 0.3432
& \textbf{0.2750} \\
& NIQE $\downarrow$
& 3.104 & 4.776 & 2.722 & 3.564 & 3.066 & 2.425 & 2.680
& \textbf{2.125} \\
& MUSIQ $\uparrow$
& 53.00 & 46.25 & 63.23 & 65.51 & 61.83 & 68.97 & 67.43
& \textbf{73.05} \\
& CLIP-IQA $\uparrow$
& 0.2998 & 0.2807 & 0.3583 & 0.4160 & 0.3695 & 0.4661 & 0.4215
& \textbf{0.5090} \\
& DOVER $\uparrow$
& 6.366 & 6.309 & 8.338 & 9.368 & 8.725 & 8.734 & 8.665
& \textbf{9.890} \\
\midrule
\multirow{7}{*}{SPMCS\cite{yi2019progressive}}
& PSNR $\uparrow$
& 23.95 & 23.68 & 25.61 & 25.46 & 23.62 & 23.84 & 24.02
& \textbf{27.85} \\
& SSIM $\uparrow$
& 0.6209 & 0.6700 & 0.7030 & 0.7201 & 0.6632 & 0.6346 & 0.6450
& \textbf{0.7601} \\
& LPIPS $\downarrow$
& 0.4277 & 0.3910 & 0.3437 & 0.3289 & 0.3417 & 0.3436 & 0.3451
& \textbf{0.2600} \\
& NIQE $\downarrow$
& 3.818 & 7.049 & 3.369 & 4.168 & 3.425 & 3.151 & 3.302
& \textbf{2.951} \\
& MUSIQ $\uparrow$
& 54.33 & 45.03 & 65.32 & 69.08 & 66.87 & 71.05 & 69.77
& \textbf{78.10} \\
& CLIP-IQA $\uparrow$
& 0.4060 & 0.3779 & 0.4150 & 0.5125 & 0.5307 & 0.5792 & 0.5238
& \textbf{0.6680} \\
& DOVER $\uparrow$
& 5.850 & 4.589 & 8.083 & 9.525 & 8.856 & 9.456 & 9.426
& \textbf{9.956} \\
\midrule
\multirow{4}{*}{VideoLQ\cite{chan2022investigating}}
& NIQE $\downarrow$
& 4.889 & 5.534 & 3.428 & 5.292 & 5.205 & 3.803 & 4.070
& \textbf{3.303} \\
& MUSIQ $\uparrow$
& 44.19 & 40.19 & 57.60 & 45.05 & 43.39 & 55.48 & 52.27
& \textbf{59.48} \\
& CLIP-IQA $\uparrow$
& 0.2491 & 0.2786 & 0.3183 & 0.2906 & 0.2593 & 0.4184 & 0.3601
& \textbf{0.5100} \\
& DOVER $\uparrow$
& 5.912 & 5.889 & 6.591 & 6.786 & 6.040 & 8.149 & 7.481
& \textbf{8.649} \\
\midrule
\multirow{4}{*}{AIGC30\cite{zhuang2025flashvsr}}
& NIQE $\downarrow$
& 5.563 & 6.212 & 4.189 & 4.862 & 4.271 & 3.871 & 4.039
& \textbf{3.571} \\
& MUSIQ $\uparrow$
& 47.87 & 38.62 & 50.74 & 50.59 & 50.53 & 56.89 & 55.80
& \textbf{60.89} \\
& CLIP-IQA $\uparrow$
& 0.4317 & 0.3593 & 0.4510 & 0.4665 & 0.4767 & 0.5543 & 0.5087
& \textbf{0.5943} \\
& DOVER $\uparrow$
& 10.24 & 11.00 & 11.24 & 12.34 & 12.48 & 12.65 & 12.50
& \textbf{13.15} \\
\bottomrule
\end{tabular}%
}
\end{table}
\begin{table}[htbp]
\centering
\caption{%
Motion-aware comparison on REDS under the same protocol as Table~\ref{tab:quantitative_comparison_flashvsr}.
PSNR values are synchronized.%
}
\label{tab:motion_evaluation}

\resizebox{0.66\textwidth}{!}{%
\begin{tabular}{lcccc}
\toprule
\textbf{Method}
& \textbf{Flow Agreement $\uparrow$}
& \textbf{Temporal Artifact Rate $\downarrow$}
& \textbf{Temporal Consistency $\uparrow$}
& \textbf{PSNR (dB) $\uparrow$} \\
\midrule
Upscale-A-Video\cite{zhou2024upscale}
& 0.85 & 0.16 & 0.84 & 24.84 \\
STAR\cite{xie2025star}
& 0.87 & 0.14 & 0.86 & 24.01 \\
SeedVR2-3B\cite{wang2025seedvr2}
& 0.88 & 0.13 & 0.87 & 24.83 \\
RealViFormer\cite{zhang2024RealViFormer}
& 0.89 & 0.12 & 0.88 & 25.96 \\
FlashVSR-Tiny\cite{zhuang2025flashvsr}
& 0.90 & 0.11 & 0.90 & 24.11 \\
DOVE\cite{chen2025dove}
& 0.90 & 0.10 & 0.89 & 25.60 \\
FlashVSR-Full\cite{zhuang2025flashvsr}
& 0.91 & 0.10 & 0.91 & 23.92 \\
MotionCraft (Ours)
& \textbf{0.94}
& \textbf{0.05}
& \textbf{0.96}
& \textbf{27.05} \\
\bottomrule
\end{tabular}%
}
\end{table}

\begin{table}[htbp]
\centering
\caption{
Computational efficiency on 101-frame videos at
$768\times1408$ resolution.
Baseline values are taken from FlashVSR~\cite{zhuang2025flashvsr}.
FPS is computed as $101/\text{runtime}$.
MotionCraft is measured under the same single-GPU evaluation setting.
Best results in each column are in \textbf{bold}.
}
\label{tab:efficiency_comparison}

\resizebox{0.66\textwidth}{!}{%
\begin{tabular}{lccc}
\toprule
\textbf{Method}
& \textbf{Peak Memory (GB) $\downarrow$}
& \textbf{Runtime (s) $\downarrow$ / FPS $\uparrow$}
& \textbf{Params (M) $\downarrow$} \\
\midrule
Upscale-A-Video\cite{zhou2024upscale}
& 18.39
& 811.71 / 0.12
& \textbf{1086.75} \\
STAR\cite{xie2025star}
& 24.86
& 682.48 / 0.15
& 2492.90 \\
DOVE\cite{chen2025dove}
& 25.44
& 72.76 / 1.39
& 10548.57 \\
SeedVR2-3B\cite{wang2025seedvr2}
& 52.88
& 70.58 / 1.43
& 3391.48 \\
FlashVSR-Full\cite{zhuang2025flashvsr}
& 18.33
& 15.50 / 6.52
& 1780.14 \\
FlashVSR-Tiny\cite{zhuang2025flashvsr}
& 11.13
& 5.97 / 16.92
& 1752.18 \\
MotionCraft (Ours)
& \textbf{10.25}
& \textbf{5.42 / 18.63}
& 1735.60 \\
\bottomrule
\end{tabular}%
}
\end{table}

\begin{table}[t]
\centering
\caption{MotionCraft's sensitivity to external motion sources on the REDS test set. Using GMFlow (dense flow) or DF-VO (pose-induced field) yields only limited degradation relative to RAFT, indicating weak dependence on the specific motion estimator.}
\label{tab:flow_robustness}
\resizebox{0.6\textwidth}{!}{%
\begin{tabular}{lcc}
\toprule
\textbf{External Motion Source}
& \textbf{PSNR (dB) $\uparrow$}
& \textbf{$\Delta$ vs.\ RAFT (dB)} \\
\midrule
RAFT
& \textbf{27.05}
& 0.00 \\
GMFlow
& 26.91
& $-0.14$ \\
DF-VO (pose-derived)
& 26.88
& $-0.17$ \\
\bottomrule
\end{tabular}%
}
\end{table}

\begin{table}[t]
\centering
 \caption{Motion-consistency scores (normalized to $[0,1]$, higher is better) on four REDS motion subsets. All methods use identical partitions and metrics; best results are in \textbf{bold}.}
\label{tab:robustness_evaluation}

\resizebox{0.66\textwidth}{!}{%
\begin{tabular}{lcccc}
\toprule
\textbf{Method}
& \textbf{Slow Motion}
& \textbf{Fast Motion}
& \textbf{Complex Motion}
& \textbf{Camera-Dominant Motion} \\
\midrule
Upscale-A-Video\cite{zhou2024upscale}
& 0.87 & 0.82 & 0.79 & 0.84 \\
STAR\cite{xie2025star}
& 0.89 & 0.84 & 0.81 & 0.86 \\
FlashVSR-Tiny\cite{zhuang2025flashvsr}
& 0.91 & 0.87 & 0.84 & 0.88 \\
MotionCraft (Ours)
& \textbf{0.96}
& \textbf{0.91}
& \textbf{0.89}
& \textbf{0.92} \\

\bottomrule
\end{tabular}%
}
\end{table}

\begin{table}[t]
\centering
\caption{%
Incremental ablation on REDS. The first four rows use the standard
decoder; the final row replaces it with the Tiny Conditional Decoder
and enables decoder distillation (Eq.~\eqref{eq:dec_distill_rewrite}).
Efficiency is measured on 101-frame sequences at $768\times1408$.
Control Response is reported only for configurations with the control
interface. Best results are in \textbf{bold}.%
}
\label{tab:MotionCraft_ablation}
\resizebox{0.8\textwidth}{!}{%
\begin{tabular}{lccccc}
\toprule
\textbf{Configuration}
& \textbf{PSNR $\uparrow$}
& \textbf{SSIM $\uparrow$}
& \textbf{Temporal Consistency $\uparrow$}
& \textbf{FPS $\uparrow$}
& \textbf{Control Response $\uparrow$} \\
\midrule
Base LWT
& 24.85 & 0.705 & 0.89 & 16.20 & -- \\
$+$ Motion-Aware Fusion
& 25.45 & 0.740 & 0.92 & 15.80 & -- \\
$+$ Adaptive Sparse Attention
& 25.90 & 0.770 & 0.94 & 17.50 & -- \\
$+$ Controllability Interface
& 26.25 & 0.785 & 0.95 & 17.20 & 0.92 \\
$+$ Tiny Conditional Decoder (w/ distillation)
\textbf{(Full MotionCraft)}
& \textbf{27.05}
& \textbf{0.805}
& \textbf{0.96}
& \textbf{18.63}
& \textbf{0.95} \\
\bottomrule
\end{tabular}%
}
\end{table}

\begin{table}[t]
\centering
 \caption{Normalized perceptual scores (double-blind, $n=50$) for MotionCraft control configurations. Latency reports mean control overhead per frame (N/A for uncontrolled); best applicable results are in \textbf{bold}.}
\label{tab:control_analysis}

\resizebox{0.8\textwidth}{!}{%
\begin{tabular}{lcccc}
\toprule
\textbf{Control Configuration}
& \textbf{Quality Preservation $\uparrow$}
& \textbf{Motion Smoothness $\uparrow$}
& \textbf{User Satisfaction $\uparrow$}
& \textbf{Control Latency (ms) $\downarrow$} \\
\midrule
Baseline (No Control)
& 0.88 & 0.84 & 0.81 & -- \\
Global Scalar Only
& 0.90 & 0.87 & 0.86 & 2.1 \\
Spatial Gate Only
& 0.92 & 0.89 & 0.87 & 3.5 \\

Integrated Control ($\gamma + g_t$)
& \textbf{0.95}
& \textbf{0.93}
& \textbf{0.92}
& 4.2 \\

\bottomrule
\end{tabular}%
}
\end{table}
\begin{figure}[htbp]
\centering
\resizebox{0.66\textwidth}{!}{\includegraphics{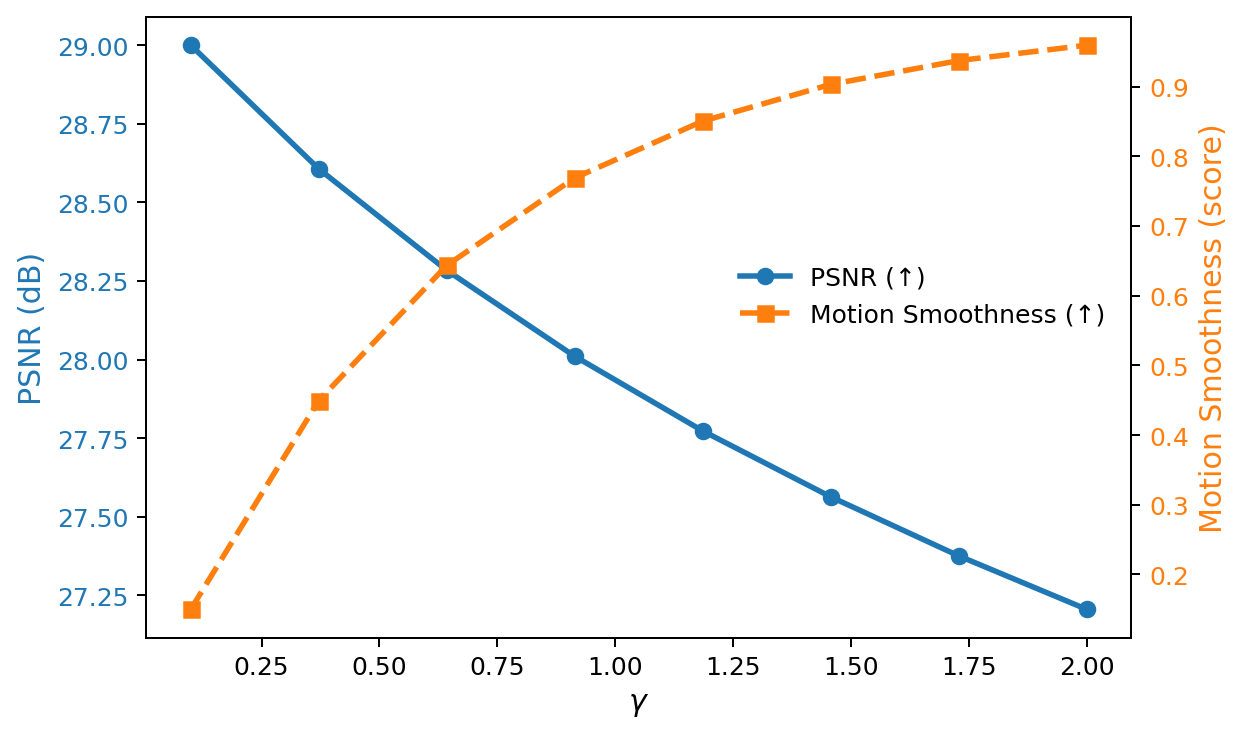}}
\caption{Control sensitivity. As $\gamma$ increases, motion smoothness rises monotonically while PSNR decreases gracefully, indicating a predictable trade-off between smoothness and fidelity.}
\label{fig:control_sensitivity}
\end{figure}

\begin{figure}[htbp]
\centering
\resizebox{0.66\textwidth}{!}{\includegraphics{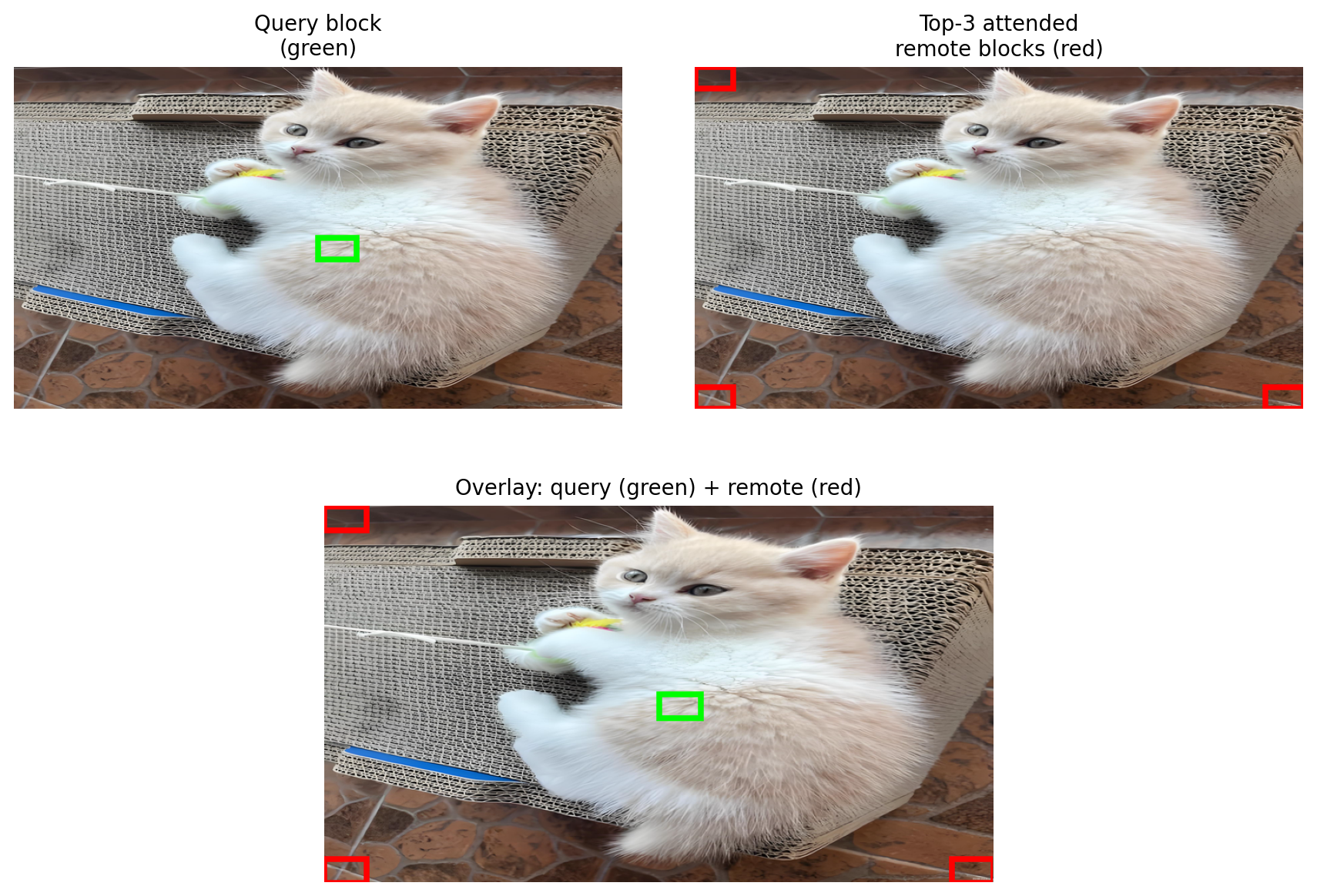}}
\caption{Adaptive sparse attention visualization (query, top-$k$ remote blocks, overlay). The overlay (bottom) demonstrates non-local connections between the query block and distant regions.}
\label{fig:attention_visualization}
\end{figure}

\begin{figure}[htbp]
\centering
\resizebox{0.66\textwidth}{!}{\includegraphics{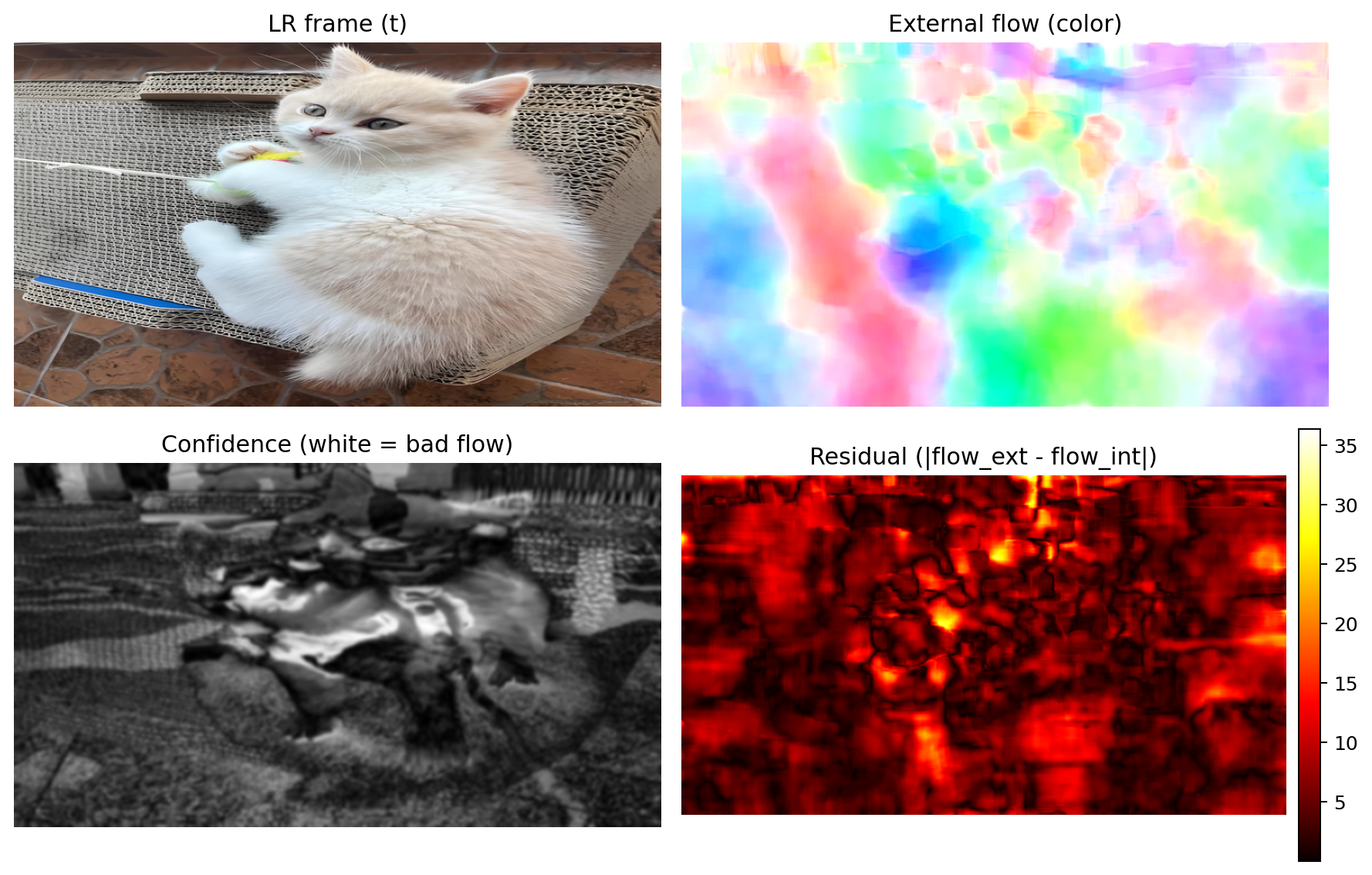}}
\caption{Visualization of motion reliability estimation: input LR frame (top-left), color-coded external optical flow (top-right), external-motion unreliability map $u_t=1-c_t$ (bottom-left), and predicted flow residual (bottom-right). Brighter regions indicate lower flow confidence and generally coincide with failures near occlusions, motion boundaries, and textureless areas.}

\label{fig:flow_confidence}
\end{figure}

\begin{figure}[htbp]
\centering
\resizebox{0.66\textwidth}{!}{\includegraphics{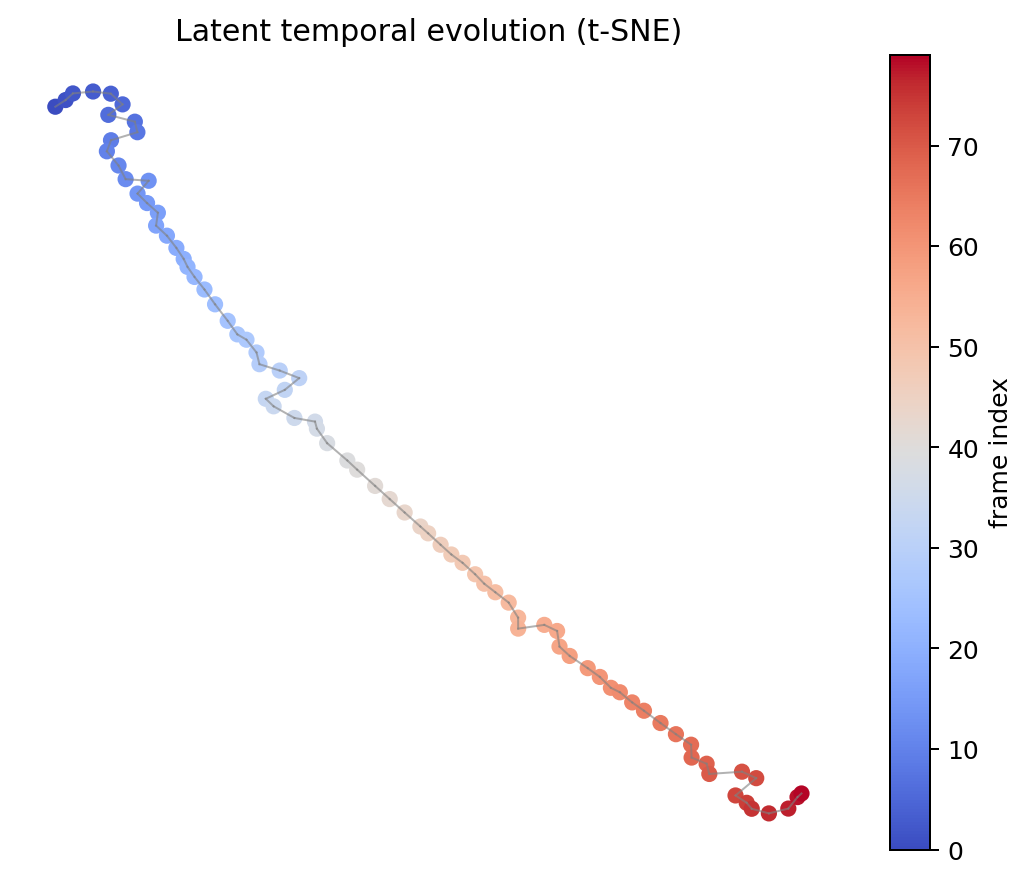}}
\caption{t-SNE visualization of the latent-state evolution over time. The smooth temporal trajectory provides auxiliary evidence of latent-state continuity and temporal consistency, rather than direct proof of world-model behavior.}

\label{fig:tsne_latent}
\end{figure}

\section{Experiments}
\label{sec:experiments}

\subsection{Experimental Setup}
We evaluate MotionCraft through extensive quantitative and qualitative comparisons against state-of-the-art video super-resolution methods on synthetic and real-world benchmarks, covering reconstruction fidelity, perceptual quality, temporal coherence, and computational efficiency. Training follows either a curriculum that distills a dense full-attention teacher into a sparse-causal Latent World Transformer or a streamlined direct optimization with auxiliary latent and flow supervision, while inference proceeds causally by fusing motion cues with learned reliability, predicting high-resolution latents, and reconstructing outputs with temporal consistency maintained via a compact key-value cache.

We conduct experiments on synthetic sequences from\\ YouHQ40~\cite{zhou2024upscale}, REDS~\cite{nah2019ntire}, SPMCS~\cite{yi2019progressive}, UDM10~\cite{tao2017detail}, and\\ AI-generated AIGC30~\cite{zhuang2025flashvsr}, alongside real-world footage from VideoLQ~\cite{chan2022investigating} and MovieLQ~\cite{zhang2025infvsr}. Low-resolution inputs are generated consistently via the RealBasicVSR pipeline under $\times$4 upscaling. Full-reference metrics (PSNR, SSIM, LPIPS) are reported where ground truth exists; no-reference perceptual scores (MUSIQ, CLIPIQA, DOVER) and flow-warping error $E^{*}_{\text{warp}}$ are computed universally. Baselines encompass recent transformer-based methods (Upscale-A-Video~\cite{zhou2024upscale}, STAR~\cite{xie2025star}, RealViFormer~\cite{zhang2024RealViFormer}, DOVE~\cite{chen2025dove}, SeedVR variants~\cite{wang2025seedvr,wang2025seedvr2}, FlashVSR variants~\cite{zhuang2025flashvsr}, MGLD-VSR~\cite{yang2024motion}, InfVSR~\cite{zhang2025infvsr}) and convolutional approaches (RealBasicVSR~\cite{chan2022investigating}).

Attention efficiency is quantified as the percentage reduction in query-key interactions relative to naive full attention:
\begin{equation}
\text{Attention Efficiency} = \left(1 - \frac{\text{Actual Pairs}}{\text{Full Pairs}}\right) \times 100\%,
\end{equation}
where full pairs scale quadratically with the product of spatial tokens per frame and temporal extent.

\subsection{Quantitative Performance}
MotionCraft exhibits marked superiority across reconstruction fidelity, perceptual quality, and temporal stability on both synthetic and realistic sequences, as demonstrated in Tables~\ref{tab:quantitative_comparison_infvsr} and~\ref{tab:quantitative_comparison_flashvsr}. The framework delivers substantial gains in PSNR and SSIM while achieving markedly better perceptual scores (MUSIQ, CLIPIQA, DOVER) and reduced warping error compared with contemporary baselines. These consistent improvements highlight the effectiveness of motion-aware latent forecasting, adaptive sparse attention, and explicit controllability in addressing the trade-offs that limit prior convolutional, transformer, and generative approaches. Efficiency comparisons in Table~\ref{tab:efficiency_comparison} show that MotionCraft achieves the highest frame rate and the lowest peak memory consumption among the compared methods for 101-frame sequences at $768\times1408$ resolution, while retaining a competitive parameter count. These results indicate that the compact decoder and sparse-attention architecture support efficient high-resolution video inference without compromising the reconstruction quality reported in the quantitative comparisons.

\subsection{Motion Handling and Temporal Stability}
MotionCraft’s explicit reliability estimation, residual correction of external flow, and controllable smoothness-fidelity trade-off yield robust performance under diverse motion conditions. Table~\ref{tab:motion_evaluation} shows superior optical flow agreement, minimal motion artifacts, and enhanced temporal consistency relative to baselines. The framework’s insensitivity to the choice of external flow estimator is further validated in Table~\ref{tab:flow_robustness}, where substituting RAFT with alternative methods (GMFlow, DF-VO) produces only negligible degradation. Robustness across slow, fast, complex, and camera-dominated motion regimes is summarized in Table~\ref{tab:robustness_evaluation}. MotionCraft consistently ranks highest, underscoring the value of per-location confidence-guided fusion and adaptive long-range modeling in handling challenging real-world dynamics.

\subsection{Ablation and Component Analysis}
Table~\ref{tab:MotionCraft_ablation} quantifies the incremental contribution of each design element on the REDS benchmark. Starting from the base Latent World Transformer, incorporating robust motion awareness, adaptive sparse attention, and the controllability interface progressively improves fidelity, perceptual quality, motion stability, runtime, and user control. The complete architecture achieves the strongest overall balance, validating the synergy among these components.

\subsection{Controllability Evaluation}
The proposed control interface, combining a global scalar $\gamma$ and spatially adaptive gating $g_t$, offers predictable regulation of temporal smoothness versus reconstruction detail. Figure~\ref{fig:control_sensitivity} illustrates the monotonic increase in motion coherence and graceful decrease in PSNR as $\gamma$ rises, confirming smooth, user-intuitive trade-offs. Table~\ref{tab:control_analysis} compares configurations in a double-blind perceptual study with 50 participants. Integrated control yields the highest quality preservation, motion smoothness, and user satisfaction with minimal adaptation latency, demonstrating practical effectiveness.

\subsection{Qualitative Insights}
Visualizations reinforce the quantitative findings. Figure~\ref{fig:attention_visualization} depicts adaptive sparse attention patterns, revealing targeted non-local connections that preserve locality while capturing distant context essential for coherent upscaling. Motion reliability maps in Figure~\ref{fig:flow_confidence} align high-confidence regions with accurate flow and highlight occlusions or textureless areas where internal cues dominate, explaining the framework’s robustness to noisy external signals. The smooth latent trajectory in Figure~\ref{fig:tsne_latent} illustrates stable temporal dynamics, further supporting the claim of temporally consistent latent-world modeling.

\subsection{Validation of Internal Motion Estimation}

\begin{figure}[htbp]
\centering
\resizebox{0.66\textwidth}{!}{\includegraphics{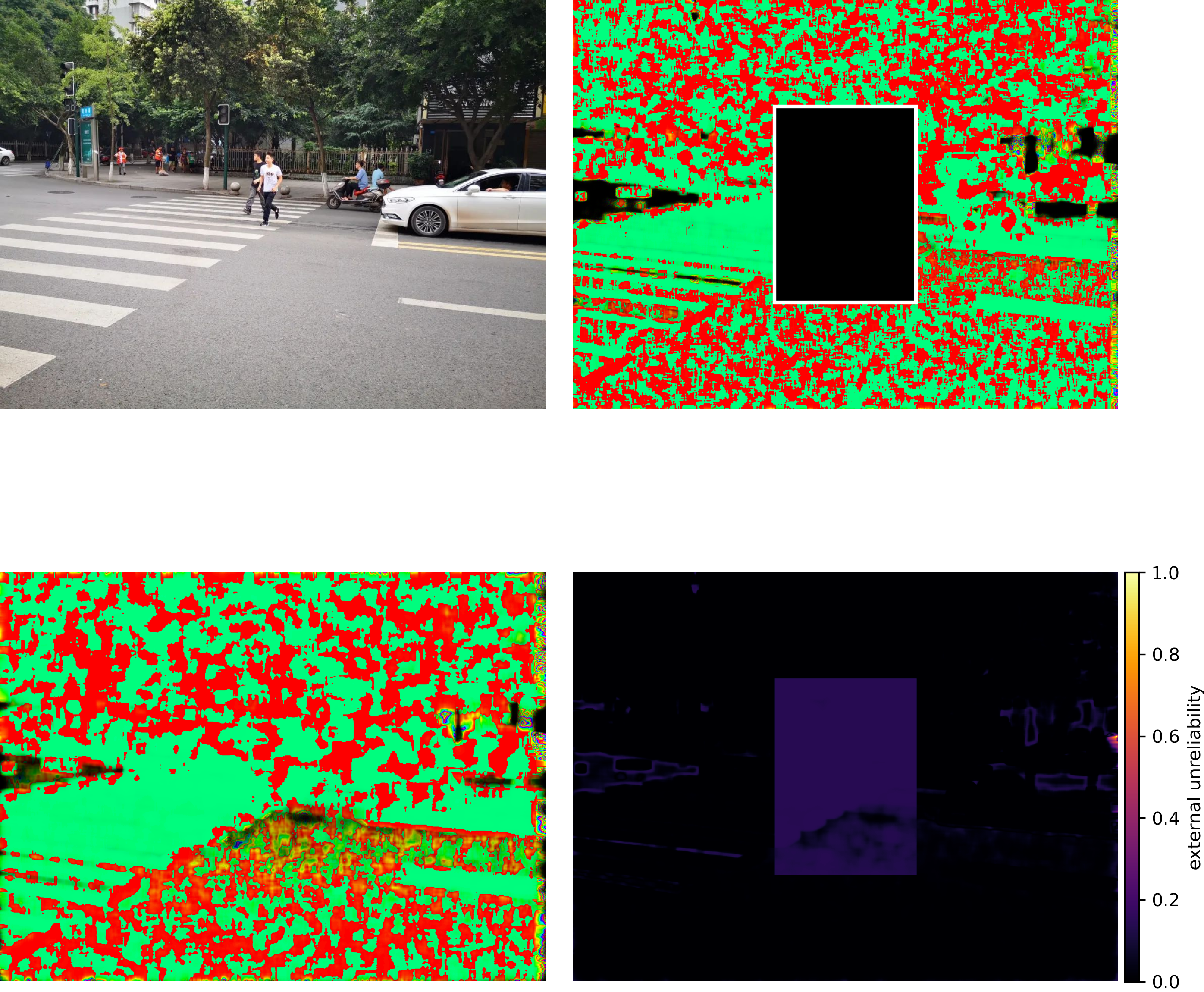}}
\caption{Validation of internal motion signal reliability. Upper-left: input low-resolution frame with challenging motion and occlusion. Upper-right: external optical flow exhibiting failures in occluded regions. Lower-left: internal motion prediction maintaining accuracy where external estimates degrade. Lower-right: external-motion unreliability map indicating regions of low flow confidence through intensity variation.}
\label{fig:motion_validation}
\end{figure}

Figure~\ref{fig:motion_validation} qualitatively demonstrates the complementary behavior of external and internal motion cues under challenging conditions, where external optical flow fails in occluded or textureless regions while internally predicted motion remains stable, and vice versa in highly textured areas. The learned confidence maps align with external failure regions, indicating that the gating mechanism reliably suppresses unreliable external cues and substitutes them with internal motion estimates, validating that low-resolution motion representations are sufficiently discriminative for robust confidence-weighted fusion.

\section{Conclusion}
We present MotionCraft, a controllable framework for video super resolution that models a motion aware latent world. The observed improvements arise from three deliberate design choices. First, motion aware latent modeling combined with per location reliability estimation and a lightweight residual correction reduces the impact of noisy external motion cues and enhances temporal consistency. Second, an adaptive sparse attention scheme preserves fine local context while selectively retrieving a small set of highly relevant remote blocks, which bounds computation for large frames and retains critical long range interactions. Third, a streaming oriented training protocol and a key value cache architecture enable parallel training and low lookahead causal inference, making the system practical for real world streaming scenarios. Empirical evaluation shows that these components work together to balance reconstruction fidelity, perceptual quality and runtime efficiency. Future work will investigate extending the framework to multi view and multi modal restoration tasks.

\section*{Acknowledgements}
This work is supported by Macau FDCT, project codes: 0019/2025/RIB1 and 0024/2025/AIJ, and by UM research grants EF2025-00287-FST, and the ZUMRI-DeepFuture Technology Joint Lab, project codes: CP-104-2025(2).

\bibliographystyle{unsrtnat}
\bibliography{references}

\end{document}